\documentclass{article}
\usepackage{ijcai22}

\usepackage{times}
\usepackage{soul}
\usepackage{url}
\usepackage[hidelinks]{hyperref}
\usepackage[utf8]{inputenc}
\usepackage[small]{caption}
\usepackage{graphicx}
\usepackage{amsmath}
\usepackage{amsthm}
\usepackage{booktabs}
\usepackage{algorithm}
\usepackage{algorithmic}
\usepackage{graphicx}

\usepackage{tikz}
\usepackage{comment}
\usepackage{amsmath,amssymb} 
\usepackage{color}

\usepackage{graphicx}
\usepackage{amsmath}
\usepackage{amssymb}
\usepackage{booktabs}
\usepackage{amsmath}
\usepackage{tabularx}

\title{Axially Expanded Windows for Local-Global Interaction in Vision Transformers}

\author{
Zhemin Zhang$^1$
\and
Xun Gong$^1$
\affiliations
$^1$Southwest Jiaotong University
\emails
zheminzhang@my.swjtu.edu.cn
}

\begin{document}

\maketitle

\begin{abstract}
Recently, Transformers have shown promising performance in various vision tasks. However, the high costs of global self-attention remain challenging for Transformers, especially for high-resolution vision tasks. Local self-attention runs attention computation within a limited region for the sake of efficiency, resulting in insufficient context modeling as their receptive fields are small. When staring at an object, human eyes usually focus on a small region while taking non-attentional regions at a coarse granularity. Based on this observation, we develop an axially expanded windows (AEWin) self-attention mechanism that performs fine-grained self-attention within the local window and coarse-grained self-attention in the horizontal and vertical axes, and thus can effectively capture both short- and long-range visual dependencies. In addition, to adapt the axially expanded windows, we propose a novel shifted window strategy to capture the complete local region of the image. Incorporated with these designs, AEWin Transformer demonstrates competitive performance on common vision tasks. Speciﬁcally, it achieves 83.1\% and 84.7\% Top-1 accuracy with the model size of 22M and 77M, respectively, testing on ImageNet-1K, outperforming the previous Vision Transformer backbones. On ADE20K semantic segmentation and COCO object detection, our AEWin Transformer backbone outperforms the recent state-of-the-art CSWin Transformer.
\end{abstract}

\section{Introduction}

Modeling in computer vision has long been dominated by convolutional neural networks (CNNs). Recently, transformer models in the field of natural language processing (NLP) \cite{DBLP:journals/corr/abs-1810-04805,NIPS2017-3f5ee243,10.1145/3437963.3441667} have attracted great interest from computer vision (CV) researchers. The Vision Transformer (ViT) \cite{DBLP:journals/corr/abs-2010-11929} model and its variants have gained state-of-the-art results on many core vision tasks \cite{Zhao2020CVPR,pmlr-v139-touvron21a}. The original ViT, inherited from NLP, ﬁrst splits an input image into patches, while equipped with a trainable class (CLS) token that is appended to the input patch tokens. Then, patches are treated in the same way as tokens in NLP applications, using self-attention layers for global information communication, and finally using the output CLS token for prediction. Recent work \cite{DBLP:journals/corr/abs-2010-11929,Liu-2021-ICCV} shows that ViT outperforms state-of-the-art convolutional networks \cite{Huang-2018-CVPR} on large-scale datasets. However, when trained on smaller datasets, ViT usually underperforms its counterparts based on convolutional layers.

The original ViT lacks inductive bias, such as locality and translation equivariance, which leads to overﬁtting and data inefficient usage. To improve data efficiency, numerous eﬀorts have studied how to introduce the locality of the CNN model into the ViT to improve its scalability \cite{NEURIPS2021-4e0928de,DBLP:journals/corr/abs-2107-00641}. These methods typically re-introduce hierarchical architectures to compensate for the loss of non-locality, such as the Swin Transformer \cite{Liu-2021-ICCV}. 

\begin{figure}[t]
  \centering
   \includegraphics[width=1.0\linewidth]{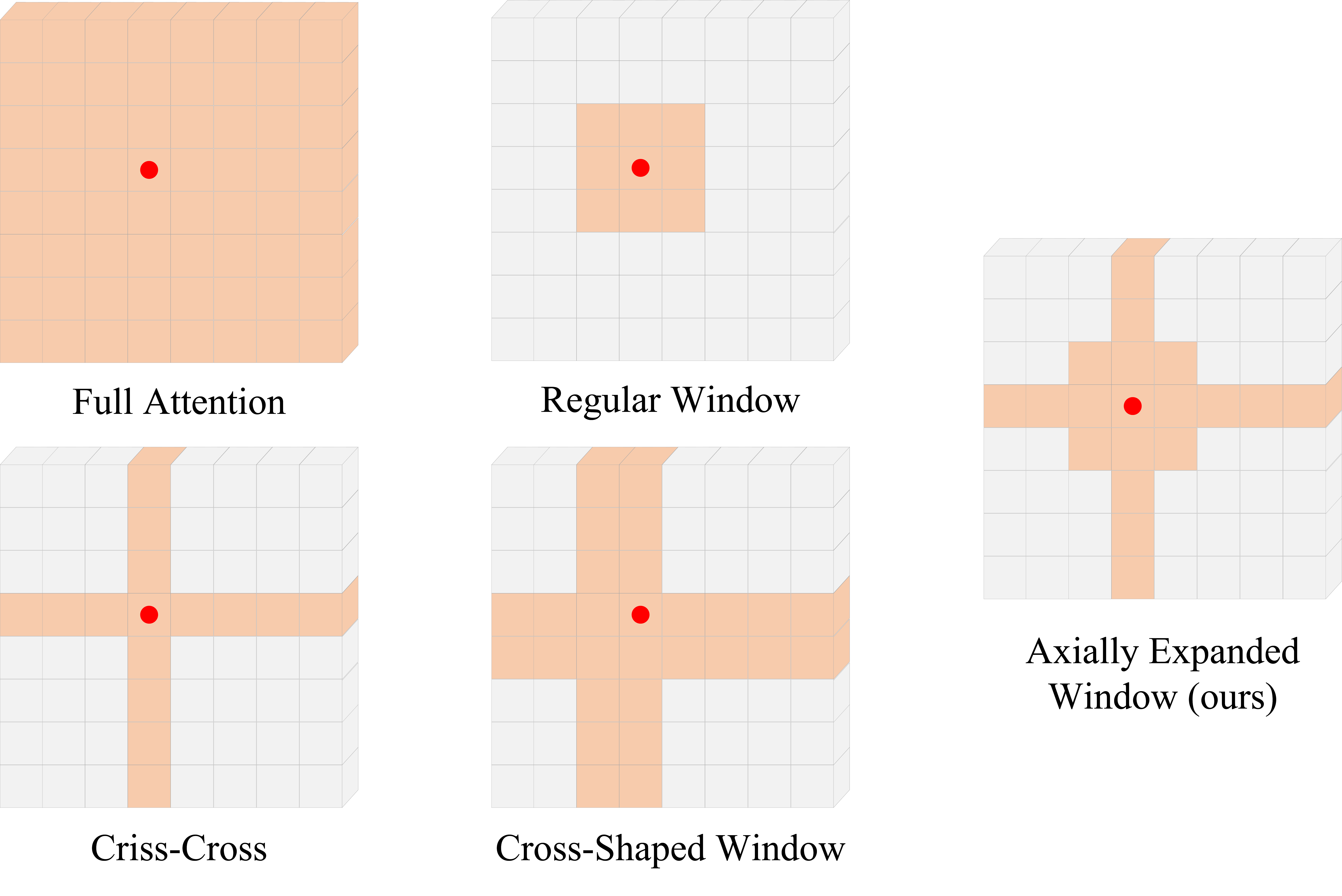}

   \caption{Illustration of different self-attention mechanisms in Transformers. Our AEWin is different in two aspects. First, we split multi-heads into three groups and perform self-attention simultaneously in the local window, horizontal and vertical axes. Second, we set different token lengths for window attention and axial attention to achieve fine-grained local and coarse-grained global interactions, which can achieve a better trade-off between computation cost and capability.}
   \label{DifferentAttentionMechanisms-flabel}
\end{figure}

Local self-attention and hierarchical ViT (LSAH-ViT) has been demonstrated to solve data inefﬁciency and alleviate model overfitting. However, LSAH-ViT uses window-based attention at shallow layers, losing the non-locality of original ViT, which leads to LSAH-ViT having limited model capacity and henceforth scales unfavorably on larger datasets such as ImageNet-21K \cite{NEURIPS2021-20568692}. To bridge the connection between windows, previous LSAH-ViT works propose specialized designs such as the “haloing operation” \cite{Vaswani-2021-CVPR} and “shifted window” \cite{Liu-2021-ICCV}. These approaches often need complex architectures, and their receptive field is increased quite slowly and requires stacking many blocks to achieve global self-attention.

\begin{figure}[t]
  \centering
   \includegraphics[width=1.0\linewidth]{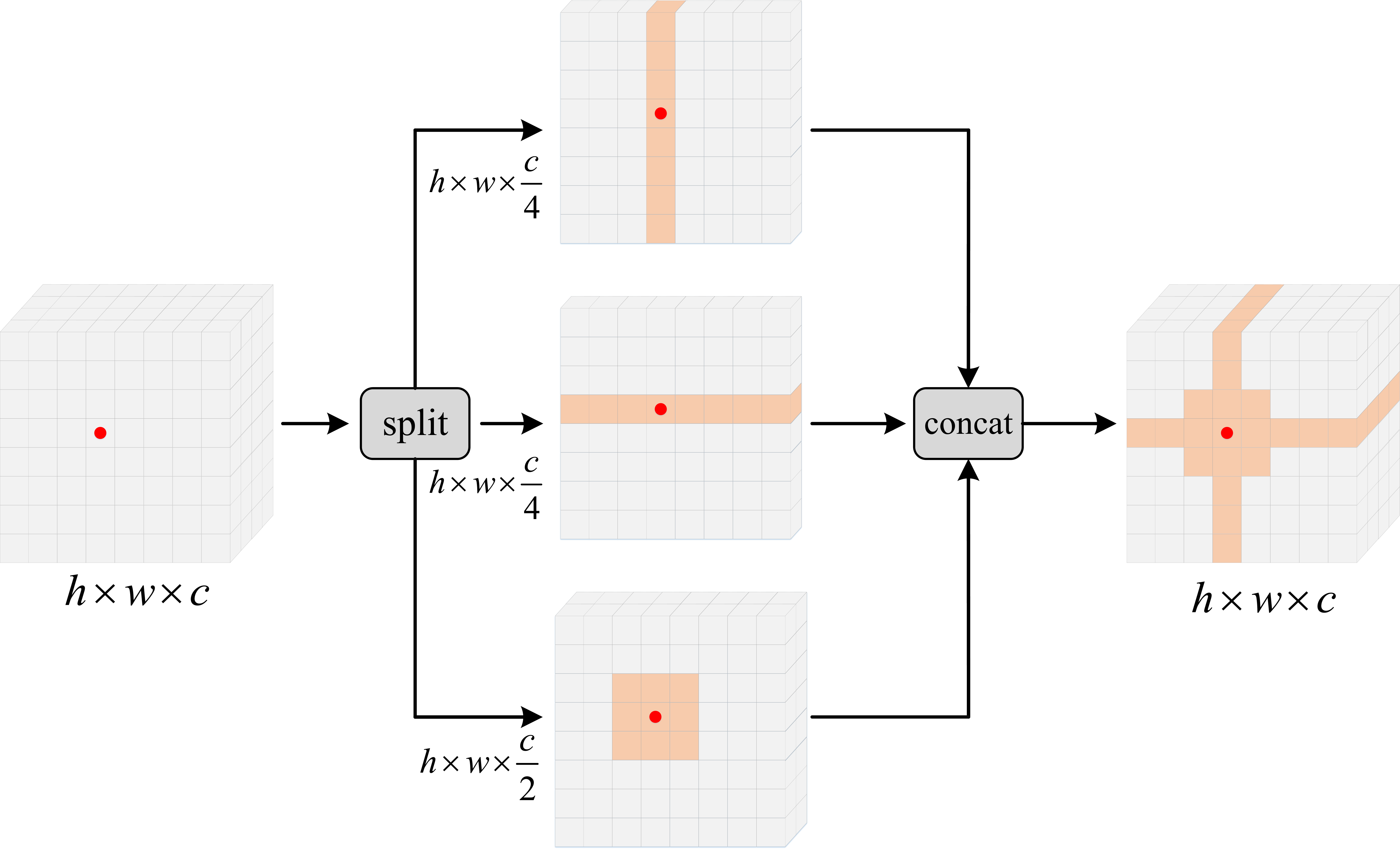}
   \caption{A parallel implementation of AEWin. It is worthwhile to note that the token length of axial attention is only half of that of windowed attention, so as to set different granularity for local and global.}
   \label{SplitGroup-flabel}
\end{figure}

When staring at an object, human eyes usually focus on a local region while attending to non-attentional regions at coarse granularity. Based on this observation, we propose an Axially Expanded Windows (AEWin) self-attention, which is illustrated in Figure \ref{DifferentAttentionMechanisms-flabel} and compared with existing self-attention mechanisms. Considering that the visual dependencies between nearby regions are usually stronger than those far away, we perform the fine-grained self-attention within the local window and coarse-grained attention on the horizontal and vertical axes. We split the multi-heads into three parallel groups and the number of heads in the first two groups is half of that in the final group, the first two groups are used for self-attention on the horizontal and vertical axes, respectively, and the final group is used for self-attention within the local window. It is worth noting that with AEWin self-attention mechanism, the self-attention in the local window, horizontal axis, and vertical axis are calculated in parallel, and this parallel strategy does not introduce extra computation costs. As shown in Figure \ref{SplitGroup-flabel}, the feature map focuses on its closest surroundings with long tokens and the surrounding on its horizontal and vertical axes with short tokens to capture coarse-grained visual dependencies. Therefore, it can capture both short- and long-range visual dependencies efficiently. Benefit from the fine-grained window self-attention and coarse-grained axial self-attention, the proposed AEWin self-attention can better balance performance and computational cost compared to existing local self-attention mechanisms shown in Figure \ref{DifferentAttentionMechanisms-flabel}.

\begin{figure}[t]
  \centering
   \includegraphics[width=1.0\linewidth]{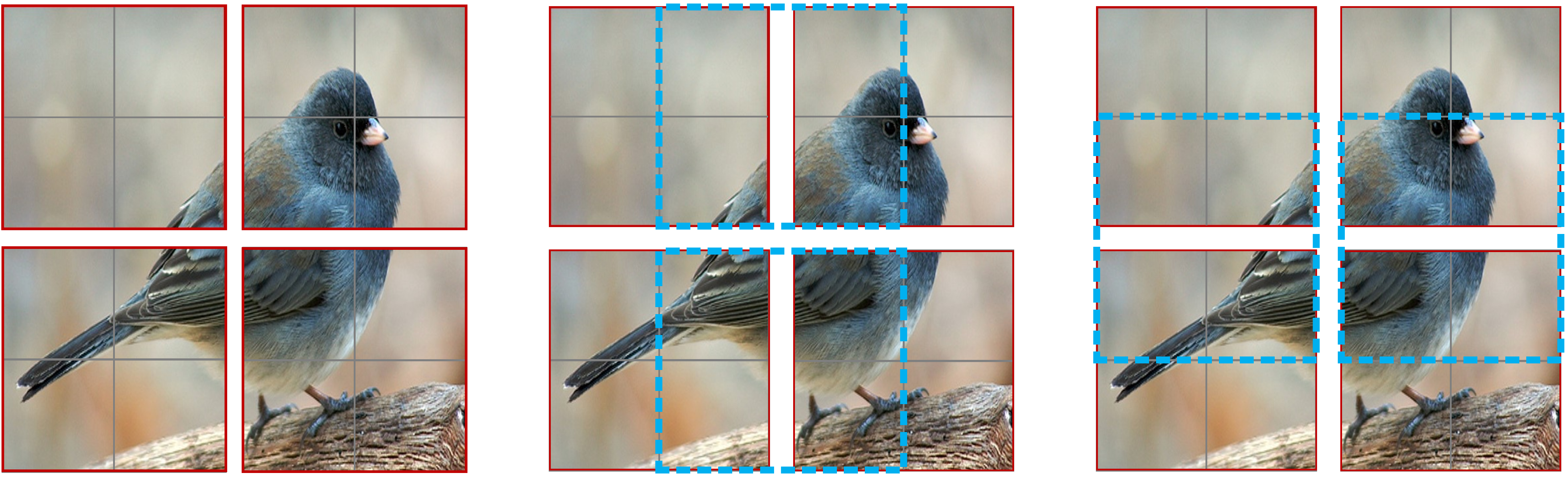}
   \caption{An illustration of the drawback of the local window attention. There is no information communication between the tokens in the window partitioned by the blue dotted line.}
   \label{BlueNoComuWin-flabel}
\end{figure}

The existing local window attention has a crucial issue for our AEWin, it cannot cover all local windows. As shown in Figure \ref{BlueNoComuWin-flabel}, there is no information communication inside the local window partitioned by the blue dotted line. Inspired by Swin Transformer \cite{Liu-2021-ICCV}, we propose a new shifted window strategy for AEWin. We will demonstrate this shifted window strategy in detail in Section 3.2.

Based on the proposed AEWin self-attention, we design a general vision transformer backbone with a hierarchical architecture, named AEWin Transformer. Our tiny variant AEWin-T achieves 83.1\% Top-1 accuracy on ImageNet-1K without any extra training data.

\begin{figure*}[t]
\centering
\includegraphics[width=1.0\linewidth]{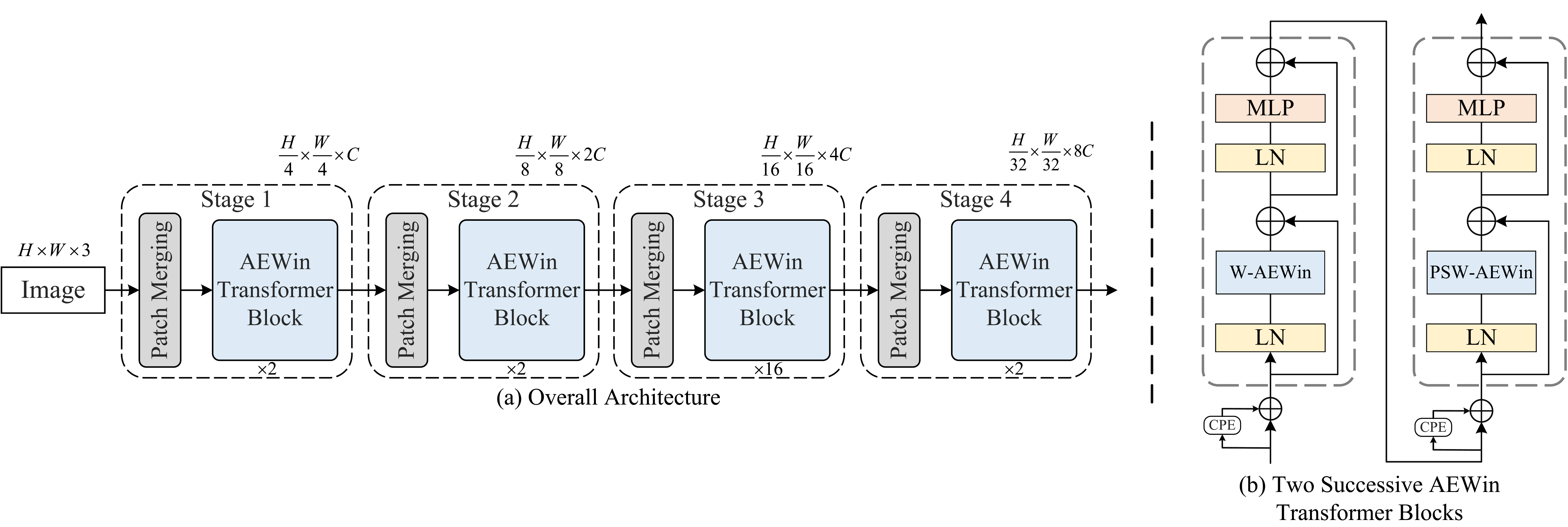} 
\caption{(a) The overall architecture of our AEWin Transformer. (b) The composition of each block. }
\label{OverallArchitecture-flabel}
\end{figure*}

\section{Related Work}

Transformers were proposed by Vaswani et al. \cite{NIPS2017-3f5ee243} for machine translation, and have since become the state-of-the-art method in many NLP tasks. Recently, ViT \cite{DBLP:journals/corr/abs-2010-11929} demonstrates that pure Transformer-based architectures can also achieve very competitive results. One challenge for vision transformer-based models is data efficiency. Although ViT \cite{DBLP:journals/corr/abs-2010-11929} can perform better than convolutional networks with hundreds of millions of images for pre-training, such a data requirement is difficult to meet in many cases. 

To improve data efficiency, many recent works have focused on introducing the locality and hierarchical structure of convolutional neural networks into ViT, proposing a series of local and hierarchical ViT. The Swin Transformer \cite{Liu-2021-ICCV} pays attention on shifted windows in a hierarchical architecture. Nested ViT \cite{zhang2022nested} proposes a block aggregation module, which can more easily achieve cross-block non-local information communication. Focal ViT \cite{DBLP:journals/corr/abs-2107-00641} presents focal self-attention, each token attends its closest surrounding tokens at ﬁne granularity and the tokens far away at coarse granularity, which can effectively capture both short- and long-range visual dependencies.

Based on the local window, a series of local self-attentions with different shapes are proposed in subsequent work. Axial self-attention \cite{DBLP:journals/corr/abs-1912-12180} and criss-cross attention \cite{Huang-2019-ICCV} achieve longer-range dependencies in horizontal and vertical directions respectively by performing self-attention in each single row or column of the feature map. CSWin \cite{Dong-2022-CVPR} proposed a cross-shaped window self-attention region, including multiple rows and columns. Pale Transformer \cite{DBLP:journals/corr/abs-2112-14000} proposes a Pale-Shaped self-Attention, which performs self-attention within a pale-shaped region to capture richer contextual information. The performance of the above attention mechanisms are either limited by the restricted window size or has a high computation cost, which cannot achieve a better trade-off between computation cost and global-local interaction.

This paper proposes a new hierarchical vision Transformer backbone by introducing axially expanded windows self-attention. Focal ViT \cite{DBLP:journals/corr/abs-2107-00641} and CSWin \cite{Dong-2022-CVPR} are the most related works with our AEWin. Compared to them, AEWin allows a better trade-off between computation cost and global-local interaction.

\section{Method}

\subsection{Axially Expanded Windows Self-Attention}

LSAH-ViT uses window-based attention at shallow layers, lacking the non-locality of original ViT, which leads to LSAH-ViT having limited model capacity and henceforth scales unfavorably on larger datasets. Existing works use specialized designs, such as the “haloing operation” \cite{Vaswani-2021-CVPR} and “shifted window” \cite{Liu-2021-ICCV}, to communicate information between windows. These approaches often need complex architectures, and their receptive field is increased quite slowly and requires stacking many blocks to achieve global self-attention. For capturing dependencies varied from short-range to long-range, inspired by the human vision system, we propose Axially Expanded Windows Self-Attention (AEWin-Attention), which performs fine-grained self-attention within the local window and coarse-grained self-attention on the horizontal and vertical axes.

\noindent \textbf{Axially Expanded Windows.} According to the multi-head self-attention mechanism, the input feature $X\in {{R}^{(H\times W)\times C}}$ will be first linearly projected to $K$ heads, and then each head will perform local self-attention within the window or horizontal axis or vertical axis.

For horizontal axial self-attention, $X$ is evenly split into non-overlapping horizontal stripes $[{{X}^{1}},\cdots ,{{X}^{H}}]$, and each stripe contains $1\times W$ tokens. Formally, suppose the projected queries, keys and values of the ${{k}^{th}}$ head all have dimension ${{d}_{k}}$,  then the output of the horizontal axis self-attention for ${{k}^{th}}$ head is deﬁned as:

\begin{equation}
\begin{aligned}
  & X=[{{X}^{1}},{{X}^{2}},\cdots ,{{X}^{H}}], \\ 
 & Y_{k}^{i}=\text{MSA}({{X}^{i}}W_{k}^{Q},{{X}^{i}}W_{k}^{K},{{X}^{i}}W_{k}^{V}), \\ 
 & \text{H-MS}{{\text{A}}_{k}}(X)=[Y_{k}^{1},Y_{k}^{2},\cdots ,Y_{k}^{H}] \\ 
\end{aligned}
  \label{horizontalAttention-glabel}
\end{equation}
where ${{X}^{i}}\in {{R}^{(1\times W)\times C}}$, $i\in \left\{ 1,2,\cdots H \right\}$, and $\text{MSA}$ indicates the Multi-head Self-Attention. $W_{k}^{Q}\in {{R}^{C\times {{d}_{k}}}}$, $W_{k}^{K}\in {{R}^{C\times {{d}_{k}}}}$, $W_{k}^{V}\in {{R}^{C\times {{d}_{k}}}}$ represent the projection matrices of queries, keys and values for the ${{k}^{th}}$ head respectively, and ${{d}_{k}}=C/K$. The vertical axial self-attention can be similarly derived, and its output for ${{k}^{th}}$ head is denoted as $\text{V-MS}{{\text{A}}_{k}}(X)$.

For windowed the self-attention, $X$ is evenly split into non-overlapping local windows $[X_{m}^{1},\cdots ,X_{m}^{N}]$ with height and width equal to $M$, and each window contains $M\times M$ tokens. Based on the above analysis, the output of the windowed self-attention for ${{k}^{th}}$ head is defined as:
\begin{equation}
\begin{aligned}
  & {{X}_{m}}=[X_{m}^{1},X_{m}^{2},\cdots ,X_{m}^{N}], \\ 
 & Y_{k}^{i}=\text{MSA}(X_{m}^{i}W_{k}^{Q},X_{m}^{i}W_{k}^{K},X_{m}^{i}W_{k}^{V}), \\ 
 & \text{W-MS}{{\text{A}}_{k}}(X)=[Y_{k}^{1},Y_{k}^{2},\cdots ,Y_{k}^{N}] \\ 
\end{aligned}
  \label{windowAttention-glabel}
\end{equation}
where $N=(H\times W)/(M\times M)$, $M$ is set as 7 by default.

\noindent \textbf{A parallel implementation of different granularities.} We split the $K$ heads into three parallel groups, with $K/4$ heads in the first two groups and $K/2$ heads in the last group, thus building a different granularity between local and global, as shown in Figure \ref{SplitGroup-flabel}. The first group of heads perform horizontal axis self-attention, the second group of heads perform vertical axis self-attention, and the third group of heads perform local window self-attention. Finally, the outputs of these three parallel groups will be concatenated back together.
\begin{equation}
\text{hea}{{\text{d}}_{k}}=\left\{ \begin{matrix}
   \text{H-MS}{{\text{A}}_{k}}\text{(}X\text{)   }  \\
   \text{V-MS}{{\text{A}}_{k}}\text{(}X\text{)   }  \\
   \text{W-MS}{{\text{A}}_{k}}\text{(}X\text{)   }  \\
\end{matrix}\begin{array}{*{35}{l}}
   k=1,\cdots ,K/4  \\
   k=K/4+1,\cdots ,K/2  \\
   k=K/2+1,\cdots ,K  \\
\end{array} \right.
  \label{mergingtoken-glabel}
\end{equation}

\begin{equation}
\text{AEWin(}X\text{)=Concat(hea}{{\text{d}}_{1}}\text{,}\cdots \text{,hea}{{\text{d}}_{K}}\text{)}{{W}^{O}}
  \label{finalOutMlp-glabel}
\end{equation}
where ${{W}^{O}}\in {{R}^{C\times C}}$ is the commonly used projection matrix that is used to integrate the output tokens of three groups. Compared to the step-by-step implementation of axial and windowed self-attention separately, such a parallel mechanism has a lower computation complexity. It can achieve different granularities by carefully designing the number of heads in different groups.

\noindent \textbf{Complexity Analysis.} Given the input feature of size $H\times W\times C$ and window size $(M,M)$, the standard global self-attention has a computational complexity of
\begin{equation}
\Omega (\text{Global})=4HW{{C}^{2}}+2{{(HW)}^{2}}C
  \label{GlobalComplexity-glabel}
\end{equation}
however, the AEWin-Attention with a parallel implementation has a computational complexity of
\begin{equation}
\Omega (\text{AEWin})=4HW{{C}^{2}}+HWC*(\frac{1}{2}H+\frac{1}{2}W+{{M}^{2}})
  \label{AEWinComplexity-glabel}
\end{equation}
which can obviously alleviate the computation and memory burden compared with the global one, since $2HW>>(\frac{1}{2}H+\frac{1}{2}W+{{M}^{2}})$ always holds.

\begin{figure}[t]
  \centering
   \includegraphics[width=1.0\linewidth]{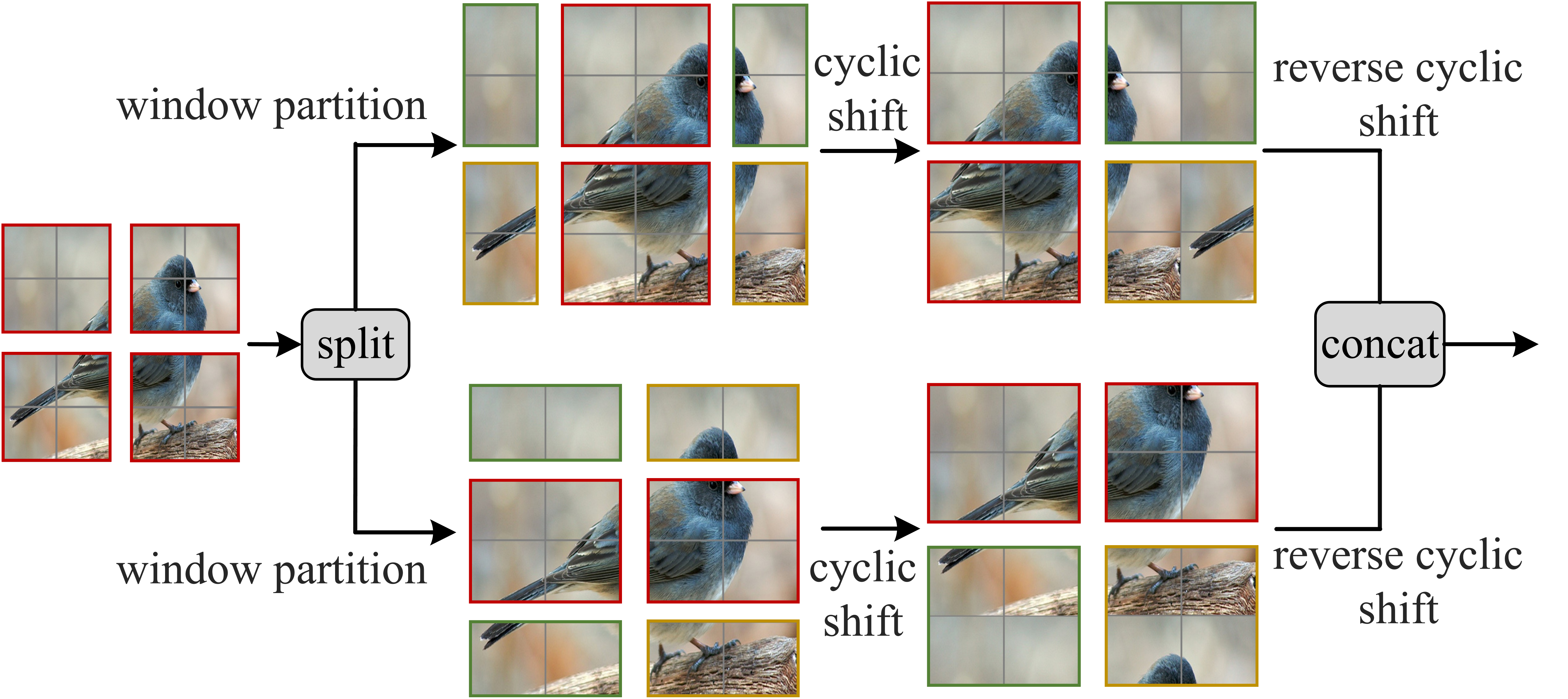}
   \caption{An illustration of the parallel shifted window approach for computing self-attention in the proposed AEWin Transformer.}
   \label{newshift-flabel}
\end{figure}

\subsection{A Parallel Implementation of Shifted Window}

The existing local window attention has a crucial issue for our AEWin, it cannot cover all local windows. As shown in Figure \ref{BlueNoComuWin-flabel}, there is no information communication inside the local window partitioned by the blue dotted line. Inspired by the Swin Transformer \cite{Liu-2021-ICCV}, we propose a new shifted window strategy that provides connections among windows while covering all local regions, significantly enhancing learning ability.

As shown in Figure \ref{newshift-flabel}, we split the heads into two parallel groups. The first and second groups displace the window right and down by $\left\lfloor \frac{M}{2} \right\rfloor$ pixels from the regularly partitioned window, respectively, and then cyclic shifting (Green and yellow boxes in Figure \ref{newshift-flabel}). Finally, the outputs of these two parallel groups will be concatenated back together. We alternate the two partition configurations in the local window calculation of consecutive AEWin Transformer blocks.

After this shift, a batched window may be composed of several sub-windows that are not adjacent in the feature map (Green and yellow boxes in Figure \ref{newshift-flabel}). As our AEWin expands the receptive field of local windows using axial attention, these non-adjacent sub-windows have information communication in the regular window partitioning strategy, so our shifted window strategy does not need to use the masking mechanism in Swin Transformer \cite{Liu-2021-ICCV}, which improves the efficiency of window attention after shifted window. The ablation study shows that our parallel shifted window strategy outperforms the Swin Transformer.

\subsection{AEWin Transformer Block}
Equipped with the above self-attention mechanism, AEWin Transformer block is formally defined as:
\begin{equation}
\begin{aligned}
 & \overset{\wedge }{\mathop{{{X}^{l}}}}\,=\text{W-AEWin}(\text{LN}({{X}^{l-1}}))+{{X}^{l-1}}, \\ 
 & {{X}^{l}}=\text{MLP}(\text{LN}(\overset{\wedge }{\mathop{{{X}^{l}}}}\,))+\overset{\wedge }{\mathop{{{X}^{l}}}}\, \\ 
 & \overset{\wedge }{\mathop{{{X}^{l+1}}}}\,=\text{PSW-AEWin}(\text{LN}({{X}^{l}}))+{{X}^{l}}, \\ 
 & {{X}^{l+1}}=\text{MLP}(\text{LN}(\overset{\wedge }{\mathop{{{X}^{l+1}}}}\,))+\overset{\wedge }{\mathop{{{X}^{l+1}}}}\, \\ 
\end{aligned}
  \label{windowAttention-glabel}
\end{equation}
where $\overset{\wedge }{\mathop{{{X}^{l}}}}\,$ and ${{X}^{l}}$ denote the output features of the $\text{AEWin}$ module and the $\text{MLP}$ module for block $l$, respectively. W-AEWin and PSW-AEWin denote AEWin self-attention using regular and parallel shifted window partitioning conﬁgurations, respectively. We use conditional positional encoding (CPE) \cite{DBLP:journals/corr/abs-2102-10882} as the positional encoding of AEWin, as shown in Figure \ref{OverallArchitecture-flabel}(b).

\subsection{Overall Architecture} 
An overview architecture of the AEWin-ViT is presented in Figure \ref{OverallArchitecture-flabel} (a), which illustrates the tiny version. AEWin-ViT consists of four hierarchical stages, like Swin-ViT \cite{Liu-2021-ICCV} to build hierarchical architecture to capture multi-scale features and alternately use shifted windows. Each stage contains a patch merging layer and several AEWin Transformer blocks. As the network gets deeper, the input features are spatially downsampled by a certain ratio through the patch merging layer. The channel dimension is expanded twice to produce a hierarchical image representation. Specifically, the spatial downsampling ratio is set to 4 in the first stage and 2 in the last three stages. The outputs of the patch merging layer are fed into the subsequent AEWin Transformer block, and the number of tokens is kept constant. Finally, we apply a global average pooling step on the output of the last block to obtain the image representation vector for the final prediction.

\begin{table}[h]
   \centering
   \caption{Detailed conﬁgurations of AEWin Transformer Variants.}
   \resizebox{\linewidth}{!}{
   \begin{tabular}{l|c|c|c|c}
      \hline 
      Models  & \#Dim & \#Blocks  &\#heads   &\#Param.                     \\
      \hline 
      AEWin-T &64   & 2,2,16,2   & 4,4,8,16       &22M     \\
      AEWin-B &96   & 2,4,24,2   & 4,8,16,32     &77M     \\
      \hline 
   \end{tabular}
   }
   \label{Variants-table}
\end{table}

\noindent \textbf{Variants.} For a fair comparison with other vision Transformers under similar settings, we designed two variants of the proposed AEWin Transformer: AEWin-T (Tiny) and AEWin-B (Base). Table \ref{Variants-table} shows the detailed conﬁgurations of all variants. They are designed by changing the block number of each stage and the base channel dimension $C$. The AEWin-T's head numbers for the four stages are 4, 4, 8, 16; the AEWin-B's head numbers are 4, 8, 16, 32.

\section{Experiments}

To show the effectiveness of the AEWin Transformer, we conduct experiments on ImageNet-1K \cite{5206848}. We then compare the performance of AEWin and state-of-the-art Transformer backbones on small datasets Caltech-256 \cite{griffin2007caltech} and Mini-ImageNet \cite{krizhevsky2012imagenet}. To further demonstrate the effectiveness and generalization of our backbone, we conduct experiments on ADE20K \cite{Zhou-2017-CVPR} for semantic segmentation, and COCO \cite{10.1007/978-3-319-10602-1-48} for object detection. Finally, we perform comprehensive ablation studies to analyze each component of the AEWin Transformer.

\subsection{Classiﬁcation on the ImageNet-1K}

\begin{table}[h]
   \centering
   \caption{Comparison of different models on ImageNet-1K.}
   \resizebox{\linewidth}{!}{
   \begin{tabular}{l|ccc|c}
      \hline 
      Method  & Image Size & Param. & FLOPs                & Top-1 acc.           \\
      \hline 
      RegNetY-4G \cite{Radosavovic-2020-CVPR} & ${{224}^{2}}$   & 21M   & 4.0G     &80.0    \\
      DeiT-S \cite{pmlr-v139-touvron21a} & ${{224}^{2}}$   & 22M   & 4.6G     &79.8    \\
      PVT-S \cite{Wang-2021-ICCV} & ${{224}^{2}}$   & 25M   & 3.8G     &79.8    \\
      Swin-T \cite{Liu-2021-ICCV} & ${{224}^{2}}$   & 29M   & 4.5G     &81.3    \\
      Focal-T \cite{DBLP:journals/corr/abs-2107-00641} & ${{224}^{2}}$   & 29M   & 4.9G     &82.2    \\
      CSWin-T \cite{Dong-2022-CVPR} & ${{224}^{2}}$   & 23M   & 4.3G     &82.7     \\
      AEWin-T (ours) & ${{224}^{2}}$   & 22M   & 4.0G     & \textbf{83.1}     \\
      \hline 
      RegNetY-16G \cite{Radosavovic-2020-CVPR} & ${{224}^{2}}$   & 84M   & 16.0G     &82.9    \\
      ViT-B  \cite{DBLP:journals/corr/abs-2010-11929} & ${{384}^{2}}$   & 86M   & 55.4G     &77.9     \\
      DeiT-B \cite{pmlr-v139-touvron21a} & ${{224}^{2}}$   & 86M   & 17.5G     &81.8    \\
      PVT-B \cite{Wang-2021-ICCV} & ${{224}^{2}}$   & 61M   & 9.8G     &81.7    \\
      Swin-B \cite{Liu-2021-ICCV} & ${{224}^{2}}$   & 88M   & 15.4G     &83.3     \\
      Focal-B \cite{DBLP:journals/corr/abs-2107-00641} & ${{224}^{2}}$   & 90M   & 16.0G     &83.8    \\
      CSWin-B \cite{Dong-2022-CVPR} & ${{224}^{2}}$   & 78M   & 15.0G     &84.2     \\
      AEWin-B (ours) & ${{224}^{2}}$   & 77M   & 14.6G     &\textbf{84.7}     \\
      \hline 
   \end{tabular}
   }
   \label{ImageNet-Top1}
\end{table}

\noindent \textbf{Implementation details.} This setting mostly follows \cite{Liu-2021-ICCV}. We use the PyTorch toolbox \cite{paszke2019pytorch} to implement all our experiments. We employ an AdamW \cite{kingma2014adam} optimizer for 300 epochs using a cosine decay learning rate scheduler and 20 epochs of linear warm-up. A batch size of 256, an initial learning rate of 0.001, and a weight decay of 0.05 are used. ViT-B/16 uses an image size 384×384 and others use 224×224. We include most of the augmentation and regularization strategies of Swin transformer\cite{Liu-2021-ICCV} in training.

\noindent \textbf{Results.} Table \ref{ImageNet-Top1} compares the performance of the proposed AEWin Transformer with the state-of-the-art CNN and Vision Transformer backbones on ImageNet-1K. Compared to ViT-B, the proposed AEWin-T model is +5.2\% better and has much lower computation complexity than ViT-B. Meanwhile, the proposed AEWin Transformer variants outperform the state-of-the-art Transformer-based backbones, and is +0.5\% higher than the most related CSWin Transformer. AEWin Transformer has the low computation complexity compared to all models in Table \ref{ImageNet-Top1}. For example, AEWin-T achieves 83.1\% Top-1 accuracy with only 4.0G FLOPs. And for the base model setting, our AEWin-B also achieves the best performance.

\begin{table}[h]
   \centering
   \caption{Comparison of different models on Caltech-256.}
   \resizebox{\linewidth}{!}{
   \begin{tabular}{l|ccc|c}
      \hline 
      Method  & Image Size & Param. & FLOPs                & Top-1 acc.           \\
      \hline 
       Swin-T \cite{Liu-2021-ICCV} & ${{224}^{2}}$   & 29M   & 4.5G     &43.3    \\
       Focal-T \cite{DBLP:journals/corr/abs-2107-00641} & ${{224}^{2}}$   & 29M   & 4.9G     &45.2     \\
       CSWin-T \cite{Dong-2022-CVPR} & ${{224}^{2}}$   & 23M   & 4.3G     &47.7     \\
       AEWin-T (ours) & ${{224}^{2}}$   & 22M   & 4.0G     & \textbf{48.6}     \\
      \hline 
       ViT-B  \cite{DBLP:journals/corr/abs-2010-11929} & ${{384}^{2}}$   & 86M   & 55.4G     &37.6     \\
      Swin-B \cite{Liu-2021-ICCV} & ${{224}^{2}}$   & 88M   & 15.4G     &46.7     \\
      Focal-B \cite{DBLP:journals/corr/abs-2107-00641} & ${{224}^{2}}$   & 90M   & 16.0G     &47.1     \\
      CSWin-B \cite{Dong-2022-CVPR} & ${{224}^{2}}$   & 78M   & 15.0G     &48.5     \\
      AEWin-B (ours) & ${{224}^{2}}$   & 77M   & 14.6G     &\textbf{49.3}     \\
      \hline 
   \end{tabular}
   }
   \label{Caltech-256-Top1}
\end{table}

\begin{table*}[t]
   \centering
   \caption{Object detection and instance segmentation performance on the COCO val2017 with the Mask R-CNN framework and 1x training schedule. The models have been pre-trained on ImageNet-1K. The resolution used to calculate FLOPs is 800×1280.}
   \begin{tabular}{l|cc|cccccc}
      \hline 
      Backbone  & Params & FLOPs & $\text{AP}^{\text{box}}$ &  $\text{AP}^{\text{box}}_{\text{50}}$  &  $\text{AP}^{\text{box}}_{\text{75}}$ &  $\text{AP}^{\text{mask}}$ &  $\text{AP}^{\text{mask}}_{\text{50}}$ &  $\text{AP}^{\text{mask}}_{\text{75}}$          \\
      \hline 
      ResNet-50 \cite{He-2016-CVPR} & 44M   & 260G   &38.0  &58.6 &41.4&34.4&55.1&36.7     \\
      Twins-S \cite{NEURIPS2021-4e0928de} & 44M   & 228G   &42.7&65.6 &46.7&39.6&62.5&42.6    \\
      PVT-S \cite{Wang-2021-ICCV} & 44M   & 245G   &40.4&62.9 &43.8&37.8&60.1&40.3     \\
      Swin-T \cite{Liu-2021-ICCV} & 48M   & 264G   &43.7&66.6 &47.6&39.8&63.3&42.7    \\
      Focal-T \cite{DBLP:journals/corr/abs-2107-00641} & 49M   & 291G   &44.8&67.7 &49.2&41.0&64.7&44.2    \\
      CSWin-T \cite{Dong-2022-CVPR} & 42M   & 279G   &46.7&68.6 &51.3&42.2&65.6&45.4     \\
      AEWin-T (ours) & 41M   & 275G   &\textbf{47.3}&\textbf{69.0} &\textbf{52.1}&\textbf{42.7}&\textbf{66.2}&\textbf{46.3}          \\
      \hline 
      RegNeXt-101-64 \cite{He-2016-CVPR} & 101M   & 493G   &42.8  &63.8 &47.3&38.4&60.6&41.3     \\
      Twins-L \cite{NEURIPS2021-4e0928de} & 120M   & 474G   &45.2&67.5 &49.4&41.2&64.5&44.5    \\
      PVT-L \cite{Wang-2021-ICCV} & 81M   & 364G   &42.9&65.0 &46.6&39.5&61.9&42.5     \\
      Swin-B \cite{Liu-2021-ICCV} & 107M   & 496G   &46.9&-- &--&42.3&--&--      \\
      Focal-B \cite{DBLP:journals/corr/abs-2107-00641} & 110M   & 533G   &47.8&70.2 &52.5&43.2&67.3&46.5    \\
      CSWin-B \cite{Dong-2022-CVPR} & 97M   & 526G   &48.7&70.4 &53.9&43.9&67.8&47.3     \\
      AEWin-B (ours) & 96M & 517G &\textbf{49.1}&\textbf{70.9} &\textbf{54.1}&\textbf{44.2}&\textbf{68.2}&\textbf{47.7} \\
      \hline 
   \end{tabular}
   \label{COCO-Top1}
\end{table*}

\subsection{Classiﬁcation on Caltech-256 and Mini-ImageNet}

\noindent \textbf{Implementation details.} Follow most of the experimental settings in the above subsection and change epochs to 100.

\noindent \textbf{Results.} In Table \ref{Caltech-256-Top1} and Table \ref{Mini-ImageNet-Top1}, we compare the proposed AEWin Transformer with state-of-the-art Transformer architectures on small datasets. With the limitation of pages, we only compare with a few classical methods here. It is known that ViTs usually perform poorly on such tasks as they typically require large datasets to be trained on. The models that perform well on large-scale ImageNet do not necessarily work perform on small-scale Mini-ImageNet and Caltech-256, e.g., ViT-B has top-1 accuracy of 58.3\% and Swin-B has top-1 accuracy of 67.4\% on the Mini-ImageNet, which suggests that ViTs are more challenging to train with less data. The proposed AEWin can significantly improve the data efficiency and performs well on small datasets such as Caltech-256 and Mini-ImageNet. Compared with CSWin, it has increased by 0.8\% and 0.7\% respectively.

\begin{table}[h]
   \centering
   \caption{Comparison of different models on Mini-ImageNet.}
   \resizebox{\linewidth}{!}{
   \begin{tabular}{l|ccc|c}
      \hline 
      Method  & Image Size & Param. & FLOPs                & Top-1 acc.           \\
      \hline 
      Swin-T \cite{Liu-2021-ICCV} & ${{224}^{2}}$   & 29M   & 4.5G     &66.3    \\
      Focal-T \cite{DBLP:journals/corr/abs-2107-00641} & ${{224}^{2}}$   & 29M   & 4.9G     &67.3     \\
      CSWin-T \cite{Dong-2022-CVPR} & ${{224}^{2}}$   & 23M   & 4.3G     &66.8     \\
      AEWin-T (ours) & ${{224}^{2}}$   & 22M   & 4.0G     & \textbf{68.2}     \\
      \hline 
      ViT-B  \cite{DBLP:journals/corr/abs-2010-11929} & ${{384}^{2}}$   & 86M   & 55.4G     &58.3     \\
      Swin-B \cite{Liu-2021-ICCV} & ${{224}^{2}}$   & 88M   & 15.4G     &67.4     \\
      Focal-B \cite{DBLP:journals/corr/abs-2107-00641} & ${{224}^{2}}$   & 90M   & 16.0G     &68.5     \\
      CSWin-B \cite{Dong-2022-CVPR} & ${{224}^{2}}$   & 78M   & 15.0G     &68.4     \\
      AEWin-B (ours) & ${{224}^{2}}$   & 77M   & 14.6G     &\textbf{69.1}     \\
      \hline 
   \end{tabular}
   }
   \label{Mini-ImageNet-Top1}
\end{table}

\subsection{COCO Object Detection}

\noindent \textbf{Implementation details.} We use the Mask R-CNN \cite{He-2017-ICCV} framework to evaluate the performance of the proposed AEWin Transformer backbone on the COCO benchmark for object detection. We pretrain the backbones on the ImageNet-1K dataset and apply the ﬁnetuning strategy used in Swin Transformer \cite{Liu-2021-ICCV} on the COCO training set.

\noindent \textbf{Results.} We compare AEWin Transformer with various backbones, as shown in Table \ref{COCO-Top1}. It shows that the proposed AEWin Transformer variants clearly outperform all the CNN and Transformer counterparts. For object detection, our AEWin-T and AEWin-B achieve 47.3 and 49.1 box mAP for object detection, surpassing the previous best CSWin Transformer by +0.6 and +0.4, respectively. We also achieve similar performance gain on instance segmentation.

\begin{table}[t]
   \centering
   \caption{Comparison of the segmentation performance of different backbones on the ADE20K. All backbones are pretrained on ImageNet-1K with the size of 224 ×224. The resolution used to calculate FLOPs is 512 ×2048.}
   \resizebox{\linewidth}{!}{
   \begin{tabular}{l|cc|cc}
      \hline 
      Backbone  & Params & FLOPs & SS mIoU & MS mIoU     \\
      \hline 
      Twins-S \cite{NEURIPS2021-4e0928de} & 55M   & 905G   &46.2&47.1     \\
      Swin-T \cite{Liu-2021-ICCV} & 60M   & 945G   &44.5&45.8    \\
      Focal-T \cite{DBLP:journals/corr/abs-2107-00641} & 62M   & 998G   &45.8&47.0    \\
      CSWin-T \cite{Dong-2022-CVPR} & 60M   & 959G   &49.3&50.4     \\
      AEWin-T (ours) & 55M   & 946G   &\textbf{49.9}&\textbf{51.1}           \\
      \hline 
      Twins-L \cite{NEURIPS2021-4e0928de} & 113M   & 1164G   &48.8&50.2     \\
      Swin-B \cite{Liu-2021-ICCV} & 121M   & 1188G   &48.1&49.7     \\
      Focal-B \cite{DBLP:journals/corr/abs-2107-00641} & 126M   & 1354G   &49.0&50.5     \\
      CSWin-B \cite{Dong-2022-CVPR} & 109M   & 1222G   &50.8&51.7     \\
      AEWin-B (ours) & 105M & 1190G &\textbf{51.7}&\textbf{52.5}  \\
      \hline 
   \end{tabular}
   \label{ADE20K-Top1}
   }
\end{table}

\subsection{ADE20K Semantic Segmentation}

\noindent \textbf{Implementation details.} We further investigate the capability of AEWin Transformer for Semantic Segmentation on the ADE20K \cite{Zhou-2017-CVPR} dataset. Here we employ the widely-used UperNet \cite{Xiao-2018-ECCV} as the basic framework and followed Swin's \cite{Liu-2021-ICCV} experimental settings In Table \ref{ADE20K-Top1}, we report both the single-scale (SS) and multi-scale (MS) mIoU for better comparison.

\noindent \textbf{Results.} As shown in Table \ref{ADE20K-Top1}, our AEWin variants outperform previous state-of-the-arts under different conﬁgurations. Speciﬁcally, our AEWin-T and AEWin-B outperform the CSWin by +0.6\% and +0.9\% SS mIoU, respectively. These results show that the proposed AEWin Transformer can effectively capture the context dependencies of different distances.

\subsection{Ablation Study}

We perform ablation studies on image classification and downstream tasks for the fundamental designs of our AEWin Transformer. For a fair comparison, we use the Swin-T \cite{Liu-2021-ICCV} as the backbone for the following experiments, and only change one component for each ablation.

\begin{table}[h]
   \centering
   \caption{Comparison of different shifted window strategies.}
   \resizebox{\linewidth}{!}{
   \begin{tabular}{l|ccc}
      \hline 
        & ImageNet & COCO & ADE20k                    \\
         & top-1   & $\text{AP}^{\text{box}}$    &SS mIoU     \\
      \hline 
       Without shifting   & 80.2    &40.8  &41.6    \\
       Swin's shifted windows \cite{Liu-2021-ICCV}   & 81.3    &43.7  &44.5    \\
       Our shifted windows   & \textbf{81.8}    &\textbf{45.0}  &\textbf{45.1}    \\
      \hline 
   \end{tabular}
   }
   \label{differentShiftedwindows}
\end{table}

\noindent \textbf{Shifted windows.} Table \ref{differentShiftedwindows} reports the ablation of different shifted window strategies on the three tasks. Our parallel shifted window partitioning outperforms the shifted window strategy in the original Swin transformer by +0.5\% top-1 accuracy on ImageNet-1K, +1.3 box AP on COCO, and +0.6 mIoU on ADE20K. The results show that our parallel shifted window can build cross-window connections more efficiently.

\begin{table}[h]
   \centering
   \caption{Ablation study on the parallel shifted windows approach. w/o shifting: all self-attention modules adopt regular window partitioning, without shifting; PSW: parallel shifted windows}
   \begin{tabular}{l|ccc}
      \hline 
        & ImageNet & COCO & ADE20k                    \\
         & top-1   & $\text{AP}^{\text{box}}$    &SS mIoU     \\
      \hline 
       w/o shifting   & 82.0    &44.8  &46.1    \\
       AEWin + PSW   & \textbf{82.5}    &\textbf{45.4}  &\textbf{46.3}    \\
      \hline 
   \end{tabular}
   \label{AEWinwithparallelshiftedwindow}
\end{table}

\noindent \textbf{Parallel shifted window in AEWin.} In this subsection, we demonstrate the results of using the axially expanded windows and parallel shifted window strategy independently in AEWin transformer, as shown in Table \ref{AEWinwithparallelshiftedwindow}. The results show that the best performance can be achieved by combining the axially expanded windows and the parallel shifted window.

\noindent \textbf{Attention Mechanism Comparison.} In this subsection, we compare with existing self-attention mechanisms. As shown in Table \ref{differentMechanisms}, the proposed AEWin self-attention mechanism performs better than the existing self-attention mechanism.

\begin{table}[h]
   \centering
   \caption{Comparison of  different self-attention  mechanisms.}
   \resizebox{\linewidth}{!}{
   \begin{tabular}{l|ccc}
      \hline 
        & ImageNet & COCO & ADE20k                    \\
         & top-1   & $\text{AP}^{\text{box}}$    &SS mIoU     \\
      \hline 
       Swin's shifted windows \cite{Liu-2021-ICCV}   & 81.3    &43.7  &44.5    \\
       Spatially Sep \cite{NEURIPS2021-4e0928de}   & 81.5    &44.2  &45.8    \\
       Sequential Axial \cite{DBLP:journals/corr/abs-1912-12180}   & 81.5    &41.5  &42.9    \\
       Criss-Cross \cite{Huang-2019-ICCV}   & 81.7    &44.5  &45.9    \\
       Cross-shaped  \cite{Dong-2022-CVPR}   & 82.2    &45.0  &46.2    \\
       AEWin (ours)   & \textbf{82.5}    &\textbf{45.4}  &\textbf{46.3}    \\
      \hline 
   \end{tabular}
   }
   \label{differentMechanisms}
\end{table}

\section{Conclusions}

This work proposes a new efficient self-attention mechanism, called axially expanded windows attention (AEWin-Attention). Compared with previous local self-attention mechanisms, AEWin-Attention simulates the way humans observe an object using fine-grained attention locally and coarse-grained attention in non-attentional regions. Based on the proposed AEWin-Attention, we developed a Vision Transformer backbone called AEWin Transformer, which achieves state-of-the-art performance on ImageNet-1K image classification, COCO object detection and ADE20K semantic segmentation.

\bibliographystyle{named}
\bibliography{ijcai22}

\end{document}